\documentclass[sn-mathphys]{sn-jnl}

\jyear{2021}%

\theoremstyle{thmstyleone}%
\theoremstyle{thmstyletwo}%

\theoremstyle{thmstylethree}%

\usepackage{graphicx}%
\usepackage{amsmath,amssymb,amsfonts}%
\usepackage{amsthm}%
\usepackage{mathrsfs}%
\usepackage[title]{appendix}%
\usepackage{xcolor}%
\usepackage{multirow}%
\usepackage{textcomp}%
\usepackage{manyfoot}%
\usepackage{booktabs}%
\usepackage{algorithm}%
\usepackage{algpseudocode}%
\usepackage{listings}%
\usepackage{float}
\usepackage{graphicx}
\usepackage{graphics}
\usepackage{subfigure}
\usepackage{float}
\usepackage{epstopdf}
\usepackage{bigstrut}
\usepackage{bibentry}
\usepackage{balance}
\usepackage[misc]{ifsym}
\usepackage{comment}
\usepackage{soul}
\usepackage{picins}

\begin{document}

\title[GraphFM: Graph Factorization Machines for Feature Interaction modeling]{GraphFM: Graph Factorization Machines for Feature Interaction Modeling}


\author[1]{\fnm{Shu} \sur{Wu}}\email{shu.wu@nlpr.ia.ac.cn}
\author[2]{\fnm{Zekun} \sur{Li}}\email{zekunli@cs.ucsb.edu}
\author[1]{\fnm{Yunyue} \sur{Su}}\email{yunyue.su@ia.ac.cn}
\author[3]{\fnm{Zeyu} \sur{Cui}}\email{cuizeyu15@gmail.com}
\author[4]{\fnm{Xiaoyu} \sur{Zhang}}\email{zhxy333@gmail.com}
\author[1]{\fnm{Liang} \sur{Wang}}\email{liang.wang@nlpr.ia.ac.cn}

\affil[1]{\orgdiv{Institute of Automation}, \orgname{Chinese Academy of Science}, \orgaddress{\city{Beijing} \postcode{100190}, \country{China}}}

\affil[2]{\orgname{University of California}, \orgaddress{\city{Santa Barbara} \postcode{93106},  \country{USA}}}

\affil[3]{\orgdiv{Alibaba Group}, \orgname{DAMO Institute}, \orgaddress{\city{Beijing} \postcode{695571},  \country{China}}}

\affil[4]{\orgdiv{Institute of Information Engineering}, \orgname{Chinese Academy of Science}, \orgaddress{\city{Beijing} \postcode{695571},  \country{China}}}


\abstract{Factorization machine (FM) is a prevalent approach to modelling pairwise (second-order) feature interactions when dealing with high-dimensional sparse data. 
However, on the one hand, FMs fail to capture higher-order feature interactions suffering from combinatorial expansion. On the other hand, taking into account interactions between every pair of features may introduce noise and degrade the prediction accuracy. 
To solve these problems, we propose a novel approach, the graph factorization machine (GraphFM), which naturally represents features in the graph structure.
In particular, we design a mechanism to select beneficial feature interactions and formulate them as edges between features.
Then the proposed model, which integrates the interaction function of the FM into the feature aggregation strategy of the graph neural network (GNN), can model arbitrary-order feature interactions on graph-structured features by stacking layers.
Experimental results on several real-world datasets demonstrate the rationality and effectiveness of our proposed approach. The code and data are available at  \href{https://github.com/CRIPAC-DIG/GraphCTR}{https://github.com/CRIPAC-DIG/GraphCTR}.}





\maketitle

\section{Introduction}\label{sect:intro}

Predictive analytics is a fundamental task in  machine learning (ML) and data mining (DM), which involves using input features to predict an output target, such as a real value for regression or categorical labels for classification.
This is particularly important for web applications, such as online advertising and recommender systems \cite{he2017neural1,he2017neural2,zhang2022novel}.
Distinct from continuous features which can be naturally found in images and audio, the features for web applications are mostly sparse and categorical.
To accurately perform predictive analytics on these types of features, it is important to consider the interactions between them. As an example, consider a scenario in which we want to predict users' preferences for movies based on five categorical variables: (1) \emph{Language = \{English, Chinese, Japanese, ... \}}, (2) \emph{Genre = \{action, fiction, ... \}}, (3) \emph{Director = \{Ang Lee, Christopher Nolan, ... \}}, (4) \emph{Stars = \{Bruce Lee, Leonardo DiCaprio, ... \}} and (5) \emph{Release Date = \{1991, 1992, ... \}}.
To capture the impact of these feature interactions, a model might consider a 3rd-order cross feature such as (\emph{Genre = fiction, Director = Christopher Nolan, Starring = Leonardo DiCaprio}) or (\emph{Language = Chinese, Genre = action, Starring = Bruce Lee}) as potentially indicating higher user preferences.

FM~\cite{rendle2010factorization,rendle2012factorization} is a popular and effective method for modelling feature interactions, that involves learning a latent vector for each one-hot encoded feature and modelling the pairwise (second-order) interactions between them through the inner product of their respective vectors. FM has been widely used in the field of recommender systems and click-through rate predictions due to its simplicity and effectiveness. However, because FM considers all feature interactions, it has two main drawbacks.

One of the main limitations of FM is that it is not able to capture higher-order feature interactions, which are interactions between three or more features. While higher-order FM (HOFM) has been proposed~\cite{rendle2010factorization,rendle2012factorization} as a way to address this issue, it suffers from high complexity due to the combination expansion of higher-order interactions. This makes HOFM difficult to use in practice. To address the limitations of FM in capturing higher-order feature interactions, several variants have been proposed that utilize deep neural networks (DNNs)~\cite{zhang2016deep,he2017neural2,cheng2016wide,guo2017deepfm}.
Factorisation machine-supported neural networks (FNNs)~\cite{zhang2016deep} apply DNNs on top of pretrained factorization machines to model high-order interactions.
Neural factorization machines (NFM)~\cite{he2017neural2} design a bi-interaction layer to learn pairwise feature interactions and apply DNNs to learn higher-order interactions.
Wide\&Deep~\cite{cheng2016wide} introduces a hybrid architecture containing both shallow and deep components to jointly learn low-order and high-order feature interactions. DeepFM~\cite{guo2017deepfm} similarly combines a shallow component with a deep component to learn both types of interactions. While these DNN-based models can effectively learn high-order feature interactions, in an implicit, bitwise manner. Consequently, they may lack the ability to provide persuasive rationales for their outputs.

In addition to not being able to effectively capture higher-order feature interactions, FM is also suboptimal because it considers the interactions between every pair of features, even if some of these interactions may not be beneficial for prediction~\cite{zhang2016understanding,su2020detecting}. These unhelpful feature interactions can introduce noise and lead to overfitting, as they do not provide useful information but make it harder to train the model. For example, in the context of predicting movie preferences, the feature interactions between \emph{Language} and \emph{Release Date} might not be relevant and, therefore, might not provide useful information for prediction. Ignoring these irrelevant feature interactions can improve model training. To solve this problem, the Attentional Decomposition Machine (AFM)~\cite{xiao2017attentional} distinguishes the importance of the factorized interaction by reweighing each cross-feature using the attentional score~\cite{bahdanau2014neural}, i.e. the influence of useless cross-features is reduced by assigning lower weights. However, it requires a predefined maximum order, which limits the potential of the model to find discriminative crossing features. Therefore, the Adaptive Factorization Network (AFN)~\cite{cheng2020adaptive} used a logarithmic neural transformation layer composed of multiple vector-wise logarithmic neurons to automatically learn the powers (i.e. the order) of features in a potentially useful combination, thereby adaptively learning cross-features and their weights from the data.

Currently, Graph Neural Networks (GNN)~\cite{kipf2016semi,hamilton2017inductive,velivckovic2017graph} have recently emerged as an effective class of models for capturing high-order relationships between nodes in a graph and have achieved state-of-the-art results on a variety of tasks such as  computer vision \cite{li2017situation}, neural language processing \cite{marcheggiani2017encoding,yao2019graph}, and recommender systems \cite{wang2019neural,wu2019session}.
At their core, GNNs learn node embeddings by iteratively aggregating features from neighboring nodes, layer by layer. This allows them to explicitly encode high-order relationships between nodes in the embeddings. GNNs have shown great potential for modelling high-order feature interactions for click-through rate prediction. Fi-GNN~\cite{li2019fi} proposed connecting each pair of features and treating the multi-field features as a fully-connected graph, using a gated graph neural network (GGNN)~\cite{li2015gated} to model feature interactions on the graph. Graph factorizer machine (GFM) \cite{xi2020graph} utilizes FM to aggregate second-order neighbour messages, and utilizes the superposition of multiple GFM layers to aggregate higher-order neighbour messages to achieve multi-order interactions from neighborhoods for recommendation. Graph-Convolved Factorization Machines (GCFM) \cite{zheng2021graph} developed the graph-convolved feature crossing (GCFC) layer to traverse all features for each input example and leveraged the features of each sample to compute the corresponding multi-feature interaction graph and propagated its influence on other features. KD-DAGFM \cite{tian2023directed} proposes a directed acyclic graph based model, that can be aligned with the DP \cite{dudzik2022graph} algorithm to improve the knowledge distillation (KD)\cite{hinton2015distilling} capability.
However, not all feature interactions are beneficial, and GNNs rely on the assumption that neighboring nodes share similar features, which may not always hold in the context of feature interaction modelling. 

In summary, when dealing with feature interactions, FM suffers intrinsic drawbacks. We thus  propose a novel model graph factorization machine (GraphFM), which takes advantage of the GNN to overcome the problems of FM for feature interaction modelling.
By treating features as nodes and feature interactions as the edges between them, the selected beneficial feature interactions can be viewed as a graph. We thus devise a novel technique to select beneficial feature interactions, which also involves inferring the graph structure. Then, we adopt an attentional aggregation strategy to aggregate these selected beneficial interactions to update the feature representations.
Specifically, to accommodate the polysemy of feature interactions in different semantic spaces, we utilize a multi-head attention mechanism~\cite{vaswani2017attention,velivckovic2017graph}.
Each layer of our proposed model produces higher-order interactions based on the existing layers; thus, the highest-order interactions are determined by layer depth. 
Since our proposed approach selects the beneficial feature interactions and models them in an explicit manner, it has high efficiency in analysing high-order feature interactions and thus provides rationales for the model outcome.
Through extensive experiments conducted on the CTR benchmark and recommender system datasets, we verify the rationality, effectiveness, and interpretability of our proposed approach. 

Overall, the main contributions of this work are threefold:
(1) We analyse the shortcomings and strengths of FM and GNN in modelling feature interactions. To solve their problems and leverage strengths, we propose a novel model GraphFM for feature interaction modelling.
(2) By treating features as nodes and their pairwise feature interactions as edges, we bridge the gap between the GNN and FM, and make it feasible to leverage the strength of the GNN to solve the FM problem.
(3) Extensive experiments are conducted on the CTR benchmark and recommender system datasets to evaluate the effectiveness and interpretability of our proposed method. We show that GraphFM can provide persuasive rationales for feature interaction modelling and prediction-making processes.

\section{Related Work}\label{sect:related}

In this work, we proposed a GNN-based approach for modelling feature interactions. We design a feature interaction selection mechanism, which can be seen as learning the graph structure by viewing the feature interactions as edges between features.
In this section, we review three lines of research that are relevant to this work: 1) techniques for learning feature interactions, 2) GNNs, and 3) graph structure learning methods.

\subsection{Feature Interaction Modelling}

Modelling feature interactions is a crucial aspect of predictive analytics and has been widely studied in the literature. FM~\cite{rendle2010factorization} is a popular method that learns pairwise feature interactions through vector inner products. Since its introduction, several variants of FM have been proposed, including field-aware factorization machine (FFM)~\cite{juan2016field} which takes into account field information and introduces field-aware embeddings, and AFM~\cite{xiao2017attentional}, which considers the weight of different second-order feature interactions. FmFM~\cite{sun2021fm2} modelled the interactions of field pairs as a matrix and utilized a kernel product to capture field interactions. However, these approaches are limited to modelling second-order interactions, which may not be sufficient in some cases.

As deep neural networks (DNNs) have proven successful in a variety of fields, researchers have begun using them to learn high-order feature interactions due to their deeper structures and nonlinear activation functions. The general approach is to concatenate the representations of different feature fields and feed them into a DNN to learn the high-order feature interactions.
Factorization-machine supported Neural Networks (FNNs)~\cite{zhang2016deep} used pretrained factorization machines to create field embeddings before applying a DNN, while product-based neural networks (PNNs)~\cite{qu2016product} model both second-order and high-order interactions through the use of a product layer between the field embedding layer and the DNN layer.
Like PNNs, neural factorization machines (NFMs)~\cite{he2017neural2} also use a separate layer to model second-order interactions, but they use a Bi-Interaction Pooling layer instead of a product layer and follow it with summation rather than concatenation. 
Other approaches to modelling second-order and high-order interactions jointly use hybrid architectures.
The Wide\&Deep \cite{cheng2016wide} and DeepFM \cite{guo2017deepfm} involve modelling low-order interactions and deep modelling high-order interactions.
However, similar to other DNN-based approaches, these models learn high-order feature interactions in an implicit, bit-wise manner and may lack transparency in their feature interaction modelling process and model outputs. As a result, some studies have attempted to learn feature interactions in an explicit fashion through the use of specifically designed networks.
Deep\&Cross~\cite{wang2017deep} introduces a CrossNet that takes the outer product of features at the bit level, while xDeepFM~\cite{lian2018xdeepfm} uses a compressed interaction network(CIN) to take the outer product at the vector level and then compresses the resulting feature maps to update the feature representations. However, xDeepFM has been found to have issues with generalizability and scalability, and it has relatively high complexity due to its consideration of all pairwise bit-level interactions. DCNV2~\cite{wang2021dcn} similarly used CIN to learn efficient explicit and implicit feature intersections, but it additionally leverages low-rank techniques to approximate feature crosses in subspaces for better performance and latency trade-offs.

More recently, several studies have attempted to use attention mechanisms to model feature interactions in a more interpretable way. HoAFM~\cite{tao2019hoafm} updates feature representations by attentively aggregating the representations of co-occurring features, while AutoInt~\cite{song2019autoint} uses a multi-head self-attention mechanism to explicitly model feature interactions.
InterHAt~\cite{li2020interpretable} is another model that uses an attentional aggregation strategy with residual connections to learn feature representations and model feature interactions. However, even with the use of attention mechanisms to account for the weight of each pair of feature interactions, aggregating all interactions together can still introduce noise and degrade the prediction accuracy. 
To address these issues, some recent studies have attempted to identify beneficial feature interactions automatically. AutoFIS~\cite{liu2020autofis} is a two-stage algorithm that uses a gate operation to search and model beneficial feature interactions, but there is a loss of information between the stages, and the modelling process is not interpretable. AFN~\cite{cheng2020adaptive} used a logarithmic neural network to adaptively learn high-order feature interactions, and the SIGN~\cite{su2020detecting} utilized mutual information to detect beneficial feature interactions and a linear aggregation strategy to model them. However, these approaches may not be expressive or interpretable enough.

\subsection{Graph Neural Networks}

A graph is a kind of data structure that reflects a set of entities (nodes) and their relationships (edges). Graph neural networks (GNNs), as deep learning architectures for graph-structured data, have attracted increasing attention.
The concept of GNNs was first proposed by~\cite{gori2005new}, and further elaborated in~\cite{scarselli2009graph}.
Currently, most of the prevailing GNN models follow the neighbourhood aggregation strategy, that is, to learn the latent node representations by aggregating the features of neighbourhoods layer by layer. 
The high-order relations between nodes can be modelled explicitly by stacking layers.
Gated graph neural networks (GGNN)~\cite{li2015gated} use GRUs~\cite{cho2014learning} to update node representations based on aggregated neighborhood feature information.
However based on graph spectral theory~\cite{bruna2013spectral}, the learning process of graph convolutional networks (GCNs)~\cite{kipf2016semi} can also be considered a mean-pooling neighborhood aggregation.
GraphSAGE~\cite{hamilton2017inductive} concatenates node features and introduces three 
mean/max/LSTM aggregators to pool neighborhood information.
Graph attention network (GAT)~\cite{velivckovic2017graph} incorporates an attention mechanism to measure the weights of neighbors when aggregating neighborhood information of a node.

Due to its strength in modelling relations on graph-structured data, the GNN has been widely applied to various applications, such as neural machine translation \cite{beck2018graph}, semantic segmentation \cite{qi20173d}, image classification \cite{marino2017more}, situation recognition \cite{li2017situation}, recommendation \cite{wu2019session,chen2020revisiting,wang2019neural}, script event prediction \cite{Zhongyang2018Constructing}, and fashion analysis \cite{cui2019dressing}.
The Fi-GNN~\cite{li2019fi} is the first attempt to exploit the GNN for feature interaction modelling.
It first proposes connecting all the feature fields; thus, the multi-field features can be treated as a fully-connected graph.
Then the GGNN~\cite{li2015gated} is utilized to model high-order feature interactions on the feature graph. KD-DAGFM~\cite{tian2023directed} uses knowledge distillation and proposes a lightweight student model, namely, directed acyclic graph FM, for learning arbitrarily explicit high-order feature interactions from teacher networks. Other graph-based work, like GFM~\cite{xi2020graph} utilized the popular factorization machine to effectively aggregate multi-order interactions in the GNN. In addition, GCFM~\cite{zheng2021graph} uses the multifilter graph-convolved feature crossing (GCFC) layer to learn the neighbor feature interactions. 

Nevertheless, the GNN was originally designed for graph classification tasks, and is based on the assumption that neighbors share similar features. As a result, the GNN inherits unnecessary and unsuitable operations for feature interaction modelling.
Our proposed model GraphFM introduces the interaction function of FM into the neighborhood aggregation strategy of the GNN to effectively capture the beneficial factorized interaction.

\section{Preliminaries}\label{sect:bg}

In this section, we first introduce the background of feature embeddings, which are fundamental for most feature interaction models based on deep learning.
To help understand our proposed model GraphFM, which is based on FMs and GNNs, we then describe these two lines of work.

\subsection{Feature Embeddings}\label{sect:emb}
In many real-world predictive tasks, such as CTR prediction, input instances consist of both sparse categorical and numerical features.
Traditionally, we represent each input instance as a sparse vector:
\begin{equation}
	\mathbf{x} = [\mathbf{x}_1,\mathbf{x}_2,...,\mathbf{x}_n],
\end{equation}
where $n$ is the number of \emph{feature fields} and $\mathbf{x}_i$ is the representation of the $i$-th feature field (aka \emph{feature}).
Since categorical features are very sparse and high-dimensional, a common way is to map them into a low-dimensional latent space. 
Specifically, a categorical feature $\mathbf{x}_i$ is mapped to dense embedding $\mathbf{e}_i \in \mathbb{R}^d$ as:
\begin{equation}
	\mathbf{e}_i = \mathbf{V}_i\mathbf{x}_i,
\end{equation}
where $\mathbf{V}_i$ denotes the embedding matrix of field $i$. 

For a numerical feature ${\bf x}_j$ which is a scalar $x_j$, we also represent it in the $d$-dimensional embedding space:
\begin{equation} 
	\mathbf{e}_j = \mathbf{v}_jx_j
\end{equation}
where $\mathbf{v}_j$ is the embedding vector for the numerical field $j$. 
Therefore, we can obtain a feature embedding matrix consisting of these feature embeddings: 
\begin{equation}
\mathbf{E}=[\mathbf{e}_1,\mathbf{e}_2,...,\mathbf{e}_n]^\top.
\end{equation}

\subsection{Factorization Machines}
The factorization machine (FM) was originally proposed for collaborative recommendation \cite{rendle2010factorization,rendle2012factorization}.
It estimates the target by modelling all interactions between each pair of features:
\begin{equation}\label{eq:fm}
\hat{y}_{\text{FM}} = \langle \mathbf{w},\mathbf{x} \rangle+\sum_{i_2>i_1}^n \langle \mathbf{e}_{i_1},\mathbf{e}_{i_2} \rangle,
\end{equation}
where $\langle\cdot,\cdot\rangle$ denotes the inner product operation. Intuitively, the first term $\langle \mathbf{w},\mathbf{x} \rangle$ is the linear regression of raw features, and the second term is the sum of all pairwise interactions, i.e., inner products of feature embeddings.

In principle, FMs can be extended to higher-order
feature combinations \cite{rendle2010factorization,rendle2012factorization}. 
Let $k \in \{2,\dots,K\}$ denote the order or degree of feature interactions considered and $\textbf{e}_{i}^{(k)}$ denote the embedding of feature $i$ for order $k$, the $K$-order higher-order FM (HOFM) can be defined as
\begin{equation}\label{eq:hofm}
\begin{aligned}
	\hat{y}_{\text{HOFM}} = \langle \mathbf{w},\mathbf{x} \rangle+\sum_{i_2>i_1}^n \langle \mathbf{e}_{i_1}^{(2)},\mathbf{e}_{i_2}^{(2)} \rangle+
	\sum_{i_3>i_2>i_1}^n \langle \mathbf{e}_{i_1}^{(3)},\mathbf{e}_{i_2}^{(3)},\mathbf{e}_{i_3}^{(3)} \rangle\\
	+\cdots+\sum_{i_K>\cdots>i_1}^n \langle \mathbf{e}_{i_1}^{(K)},\ldots,\mathbf{e}_{i_K}^{(K)} \rangle,
\end{aligned}
\end{equation}
where $\langle \textbf{e}_{i_1}^{(K)}, \dots, \textbf{e}_{i_K}^{(K)}
\rangle = \text{sum}(\textbf{e}_{i_1}^{(K)} \odot \dots \odot \textbf{e}_{i_K}^{(K)})$ (sum of
element-wise products). 
Since all the feature interactions of order up to $K$ are included, its time complexity increases exponentially, resulting in high computational complexity.
Considering that not all feature interactions are beneficial, FMs have trouble modelling higher-order feature interactions in terms of both efficiency and effectiveness.

Although several recent studies have enhanced FMs with DNNs to model higher-order feature interactions, like NFM~\cite{he2017neural2} and DeepFM~\cite{guo2017deepfm},
they model higher-order feature interactions in an implicit manner, and lack persuasive rationales for model outcomes.

\subsection{Graph Neural Networks}
Given that a graph $G=\left \{ V, E \right \}$ denotes a graph, GNNs learn the representation vectors of nodes by exploring their correlations with neighboring nodes.
The modified GNNs follow a neighborhood aggregation strategy, where we iteratively update the representation of a node by aggregating the features of its neighbors~\cite{xu2018powerful}.
After $k$ iterations of aggregation, a node's representation encodes its interaction with neighbors within $k$ hops.
The choice of aggregation strategy for GNNs is crucial. A number of architectures have been proposed. 
In this work, we adopt the attentional aggregation strategy in~\cite{velivckovic2017graph}, which will be further elaborated upon in Section~\ref{sect:model}.

\section{Graph Factorization Machine}\label{sect:model}

\subsection{Model Overview}
\begin{figure}[t]
\centering
\includegraphics[width=0.6\columnwidth]{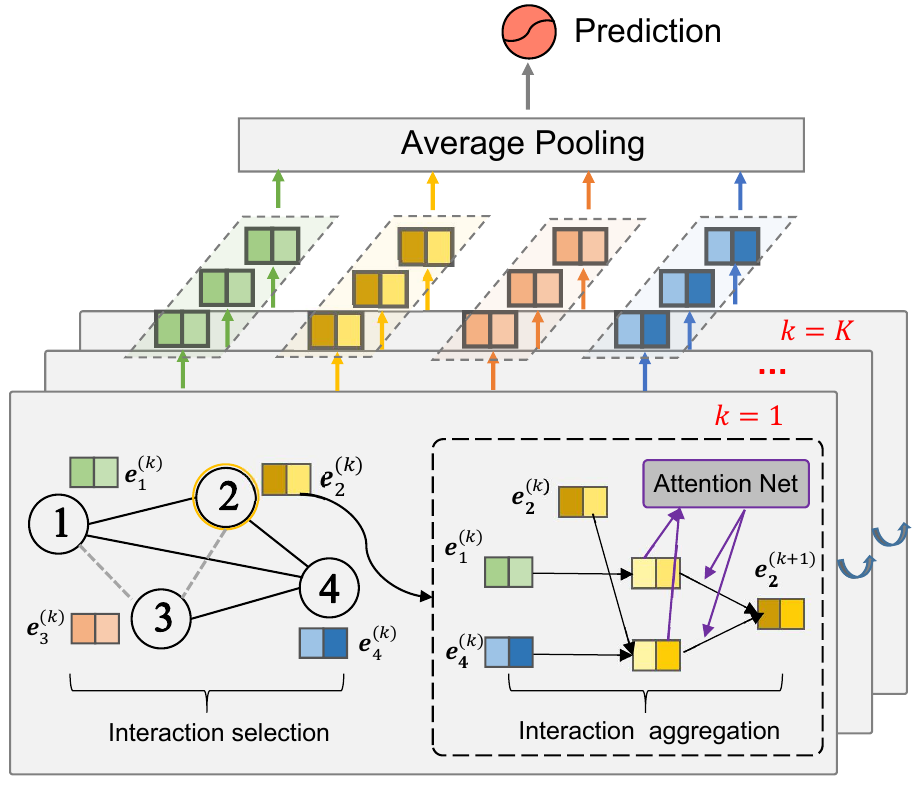}
\caption{Overview of GraphFM. The input features are modelled as a graph, where nodes are feature fields, and edges are interactions.
At each layer of GraphFM, the edges (beneficial interactions) are first selected by the \emph{interaction selection} component.
Then these selected feature interactions are aggregated via the attention network to update the feature embeddings in the \emph{interaction aggregation} component.
The learned feature embeddings at every layer are used for the final prediction jointly.}
\label{fig:framework}
\vspace{-4mm}
\end{figure}

An overview of GraphFM is shown in Fig.~\ref{fig:framework}. Each input instance (multi-field feature) is represented as a graph where the nodes are feature fields, and the edges are interactions~\cite{li2019fi,su2020detecting}.
Note that we use nodes and features, edges, and interactions interchangeably in this paper.
GraphFM updates feature representations layer by layer.
The feature embeddings described in Section~\ref{sect:emb} are taken as the initial feature embeddings of GraphFM, i.e., $\textbf{e}^{(1)}_{i}=\textbf{e}_{i}$, where $\mathbf{e}^{(k)}_{i}$ represents the updated feature embeddings at the $k$-th layer.
Since no edge information is given, we need to select the edges (beneficial interactions) by the interaction selection component first.
Then, we aggregate these selected feature interactions to update the feature embeddings in the neighborhood aggregation component.
Within each $k$-th layer, we are able to select and model only the beneficial $k$-th order feature interactions and encode these factorized interactions into feature representations.
Finally, these learned feature embedding encoded interactions of order up to $K$ are concatenated to make the final prediction.

There are two main components in each layer of GraphFM. Next, we will introduce them in detail. As we focus on describing the detailed mechanism at every single layer, we omit the layer index $k$ if not necessary.

\subsection{Interaction Selection}
To select beneficial pairwise feature interactions, we devised an interaction selection mechanism.
This can also be viewed as inferring the graph structure, which predicts the links between nodes.
However, the graph structure $G=\left \{ V, E \right \}$ is discrete, where an edge $(v_i, v_j) \in E$ linking two nodes is either present or absent. 
This makes the process non-differentiable; therefore, it cannot be directly optimized with gradient-descent-based optimization techniques.

To overcome this limitation, we replace the edge set $E$ with weighted adjacency $\mathbf{P}$, where $p_{ij}$ is interpreted as the probability of $(v_i, v_j) \in E$, which also reflects how beneficial their interaction is.  
Notably, we learn different graph structures $\mathbf{P}^{(k)}$ at each $k$-th layer.
Compared with using a fixed graph at each layer, we have more efficiency and flexibility in enumerating the beneficial higher-order feature interaction by this means. 
More specifically, by using a fixed graph structure at each layer, we can only obtain a fixed set of feature interactions. However, with the adaptive learned graph structure at each layer, the model is capable of modelling any potential feature interactions.

\subsubsection{Metric Function}~\label{sect:prob}
We aim to design a metric function between each pair of feature interactions to measure whether they are beneficial.
Formally, the weight $p_{ij}$ of an edge $(v_i, v_j)$ is computed via a metric function $f_{s}(\mathbf{e}_i, \mathbf{e}_j)$.
Here, we adopt a neural matrix factorization (NMF)~\cite{he2017neural1} based function to estimate the edge weight. 
Formally, a multi-layer perception (MLP) with one hidden layer is used to transform the element-wise product of these feature vectors to a scalar: 
\begin{equation}\label{eq:is}
f_{s}(\mathbf{e}_i, \mathbf{e}_j) = \sigma(\mathbf{W}^s_2\delta(\mathbf{W}^s_1(\mathbf{e}_i \odot \mathbf{e}_j)+\mathbf{b}^s_1)+\mathbf{b}^s_2),
\end{equation}
where $\mathbf{W}^s_1$, $\mathbf{W}^s_2$, $\mathbf{b}^s_1$, and $\mathbf{b}^s_2$ are the parameters of the MLP. $\delta(\cdot)$ and $\sigma(\cdot)$ are ReLU and sigmoid activation functions, respectively.
It should be noted that $f_{s}$ is invariant to the order of its input, i.e., $f_{s}(\mathbf{e}_i, \mathbf{e}_j) = f_{s}(\mathbf{e}_j, \mathbf{e}_i)$. Therefore, the estimated edge weights are identical for the same pair of nodes.
Such continuous modelling of the graph structure enables backpropagation of the gradients.
Since we do not have a ground-truth graph structure, the gradients come from the errors between the model output with and the target.

Intuitively, we treat the element-wise product of each pair of feature embeddings as a term and estimate its weight using the MLP.
One can also choose the Euclidean distance~\cite{kazi2020differentiable} or other distance metrics.

\subsubsection{Graph Sampling} 
From the estimated edge weighted matrix $\mathbf{P}^{(k)}$ at each layer, we then sample the beneficial feature interactions to sample the neighborhood for each feature field.
In this work, we uniformly sample a fixed-size set of neighbors. For each feature node $v_i$ at the $k$-th layer, we select $m_k$ edges according to the first $m_k$ elements of $\mathbf{P}^{(k)}[i,:]$, which can be illustrated as follows:
\begin{equation}\label{eq:sample}
\begin{aligned}
for~&i = 1, 2, \cdots, n \\
&\text{idx}_{i} = \text{argtop}_{m_k}(\mathbf{P}^{(k)}[i,:]) \\
&\mathbf{P}^{(k)}[i, -\mathbf{idx}] = 0,
\end{aligned}
\end{equation}
where $\mathbf{P}^{(k)}[i,:]$ denotes the $i$-th column of matrix $\mathbf{P}^{(k)}$ at the $k$-th layer, and $\mathbf{P}^{(k)}[i, -\text{idx}_{i}]$ contains a subset of columns of $\mathbf{P}^{(k)}$ that are not indexed by $\text{idx}_{i}$. 
$\text{argtop}_{m_k}$ is an operator that selects the $m_k$-most important nodes for query node $i$. We only retain these $m_k$ feature nodes, and the others are masked.
Thus the neighborhood set of node $v_j$ is defined as:
\begin{equation}\label{eq:neighbor}
\mathcal{N}_i^{(k)} = \left \{ v_j \mid p_{ij}^{(k)} > 0, j=1,2,\cdots,n \right \}.
\end{equation}
Practically speaking, we found that our approach could achieve high performance when $k=3$, and $m_1$ equals the number of feature fields, which means that in the first layer, we model all pairs of feature interactions.

It is worth mentioning that we also tried to set a threshold to select the edges in the graph, i.e., setting a minimum value for the edge probability of cutting edges off. However, the performance is not as good as that of using a fixed-degree graph. 
This is reasonable because the edge weights of different nodes' neighbors are at different scales. Setting a single threshold on all the nodes will lead to the situation in which the numbers of nodes' neighbors vary greatly. Some nodes will have barely any adjacent nodes after cutting off, while some may still have many.

\subsection{Interaction Aggregation}
Since we have selected the beneficial feature interactions, or in other words, learned the graph structure, we perform the interaction (neighborhood) aggregation operation to update the feature representations.

For a target feature node $v_i$, when aggregating its beneficial interactions with neighbors, we also measure the attention coefficients of each interaction.
To measure the attention coefficients, we use a learnable projection vector $\mathbf{a}$ and apply a LeakeyReLU non-linear activation function.
Formally, the coefficients are computed as:
\begin{equation}\label{eq:attention}
c_{ij} = \text{LeakyReLU}(\mathbf{a}^{\top}(\mathbf{e}_i \odot \mathbf{e}_j)).
\end{equation}
This indicates the importance of the interactions between feature $v_i$ and feature $v_j$. 

Note that we only compute $c_{ij}$ for nodes $j \in \mathcal{N}_i$, where $\mathcal{N}_i$ denotes the neighborhood of node $v_i$, which is also the set of features whose interactions with $v_i$ are beneficial.
To make coefficients easily comparable across different feature nodes, we normalized them across all choices of $j$ using a softmax function:
\begin{equation}\label{eq:softmax}
\alpha_{ij}=\frac{\text{exp}(c_{ij})}{\sum_{j'\in\mathcal{N}_i}\text{exp}(c_{ij'})}.
\end{equation}

Once the normalized attention coefficients are obtained, we compute the linear combination of these feature interactions with nonlinearity as the updated feature representations:
\begin{equation}\label{eq:update}
\mathbf{e}_{i}^{'}=\sigma\left ( \sum_{j\in\mathcal{N}_i}\alpha_{ij}p_{ij}\mathbf{W}_a(\mathbf{e}_{i} \odot \mathbf{e}_{j})\right ),
\end{equation}
where $\alpha_{ij}$ measures the attention coefficients of each feature interaction between feature $i$ and $j$, while $p_{ij}$ represents the probability of this feature interaction being beneficial.
The attention coefficient $\alpha_{ij}$ is calculated by the soft attention mechanism, while $p_{ij}$ is calculated by the hard attention mechanism. By multiplying them together, we control the information of selected feature interactions and make the parameters in the interaction selection component trainable with gradient back-propagation.

To capture the diverse polysemy of feature interactions in different semantic subspaces~\cite{li2020interpretable} and stabilize the learning process~\cite{vaswani2017attention,velivckovic2017graph}, we extend our mechanism to employ multi-head attention.
Specifically, $H$ independent attention mechanisms update Equation~\ref{eq:update}, and then these features are concatenated, resulting in the following output feature representation: 
\begin{equation}\label{eq:multi-head}
\mathbf{e}_{i}^{'}={\mid \mid}_{h=1}^H \sigma\left ( \sum_{j\in\mathcal{N}_i}\alpha_{ij}^h p_{ij}\mathbf{W}_a^h(\mathbf{e}_{i} \odot \mathbf{e}_{j})\right ),
\end{equation}
where $\parallel$ denotes the concatenation, $\alpha_{ij}^h$ is the normalized attention coefficient computed by the $h$-th attention mechanism, and $\mathbf{W}^h_a$ is the corresponding linear transformation matrix. 
One can also choose to employ average pooling to update the feature representations:
\begin{equation}
\mathbf{e}_{i}^{'} =  \sigma\left(\frac{1}{H}\sum_{h=1}^H \sum_{j\in\mathcal{N}_i}\alpha_{ij}^h  p_{ij} \mathbf{W}_a^h(\mathbf{e}_{i} \odot \mathbf{e}_{j})\right ).\end{equation}

\subsection{Prediction and Optimization}
The output of each $k$-th layer, is a set of $n$ feature representation vectors encoding feature interactions of order up to $k$, namely $\left\{\mathbf{e}^{(k)}_1, \mathbf{e}^{(k)}_2, \dots, \mathbf{e}^{(k)}_n \right \}$.
Since the representations obtained in different layers encode the interactions of different orders, they have different contributions to the final prediction. As such, we concatenate them to constitute the final representation of each feature~\cite{wang2019neural}:
\begin{equation}\label{equ:final-rep}
\mathbf{e}^{*}_i = \mathbf{e}^{(1)}_i \left \| \cdots \right \| \mathbf{e}^{(K)}_i,
\end{equation} 
Finally, we employ average pooling on the vectors of all features to obtain a graph-level output and use a projection vector $\mathbf{p}$ to make the final prediction:
\begin{equation}\label{equ:avg}
\mathbf{e}^{*} = \frac{1}{n}\sum_{i=1}^n \mathbf{e}^{*}_i,
\end{equation}
\begin{equation}\label{eq:prediction}
\hat{y} = \mathbf{p}^\top\mathbf{e}^{*}.
\end{equation}

GraphFM can be applied to various prediction tasks, including regression, classification, and ranking. 
In this work, we conduct experiments on CTR prediction, a binary classification task. We thus use log loss as the loss function:
\begin{equation}\label{eq:logloss}
	\mathcal{L} = -\frac{1}{N}\sum_{i=1}^N y_i \log\sigma(\hat{y}_i)+ (1-y_i)\log(1-\sigma(\hat{y}_i)),
\end{equation} 
where $N$ is the total number of training instances, and $\sigma$ denotes the sigmoid function.
$y_i$ and $\hat{y}_i$ denote the label of instance $i$ and the prediction of GraphFM, respectively.
The model parameters are updated using Adam\cite{kingma2014adam}.

\section{Experiments}

This section presents an empirical investigation of the performance of GraphFM on two CTR benchmark datasets and a recommender system dataset. The experimental settings are described, followed by comparisons with other state-of-the-art methods. An ablation study is also conducted to verify the importance of each component of the model and evaluate its performance under different hyperparameter settings. Finally, the question of whether GraphFM can provide interpretable explanations for its predictions is examined.

\begin{table}[t]
\centering\caption{Statistics of the evaluation datasets.}
\begin{tabular}{cccc} 
\toprule
Dataset & \#Instances & \#Fields & \#Features (sparse)  \\
\hline
Criteo & 45,840,617 & 39 & 998,960 \\
Avazu & 40,428,967 & 23 & 1,544,488 \\
MovieLens-1M & 739,012 & 7 & 3,529 \\
\bottomrule
\end{tabular}\label{tab:dataset}
\end{table}

\subsection{Experimental Settings}
Our experiments are conducted on three real-world datasets, two CTR benchmark datasets, and one recommender system dataset. Details of these datasets are listed in Table \ref{tab:dataset}.
The data preparation follows the strategy in~\cite{tian2023directed}. We randomly split all the instances at 8:1:1 for training, validation, and testing. We adopt the two most popular metrics, \textbf{AUC} and \textbf{Logloss} to measure the probability that one prediction diverges from the ground truth. 

\subsubsection{Baselines}\label{sect:baseline}
We compare GraphFM with four classes of state-of-the-art methods:(A) the linear approach that only uses individual features; (B) FM-based methods that consider second-order feature interactions; (C) DNN-based methods that model high-order feature interactions; and (D) aggregation-based methods that update features' representation and model their high-order feature interactions via an aggregation strategy.

The models associated with their respective classes are listed as follows:
\begin{itemize}
	\item \textbf{LR}~(A) refers to linear/logistics regression, which can only model linear interactions.
	\item \textbf{FM}~\cite{rendle2012factorization} ~(B) is the official FM implementation, which can only model second-order interactions.
	\item \textbf{AFM}~\cite{xiao2017attentional}~(B) is an extension of FM that considers the weights of different second-order feature interactions by using attention mechanisms.
    \item \textbf{AFN}~\cite{cheng2020adaptive} ~(B) learns arbitrary-order feature interactions adaptively from data, instead of explicitly modelling all the cross features within a fixed maximum order.
    \item \textbf{FmFM}~\cite{sun2021fm2}~(B) uses a field matrix between two feature vectors to model their interactions and learns the matrix separately for each field pair.
	\item \textbf{NFM}~\cite{he2017neural2}~(C) devises a bi-interaction layer to model second-order interactions and uses a DNN to introduce nonlinearity and model high-order interactions.
	\item \textbf{HOFM}~(C)~\cite{blondel2016higher} is the implementation of the higher-order FM~\cite{blondel2016higher}. It is a linear model.
	\item \textbf{DeepCrossing}~\cite{shan2016deep} ~(C) utilizes a DNN with residual connections to model non-linear feature interactions in an implicit manner.	
	\item \textbf{CrossNet}~\cite{wang2017deep}~(C) is the core of the Deep\&Cross model, which models feature interactions explicitly by taking the outer product of the concatenated feature vector at the bit-wise level.
	\item \textbf{xDeepFM}~\cite{lian2018xdeepfm} ~(C) takes the outer product of the stacked feature matrix at a vector-wise level to explicitly model feature interactions. ANNs can also be combined with DNNs, which model implicit and explicit interactions simultaneously.
    \item \textbf{DCNV2}~\cite{wang2021dcn} ~(C) utilizes a cross network from the DCN to learn explicit and bounded-degree cross features.
	\item \textbf{AutoInt}~\cite{song2019autoint}~(D) uses a self-attention network to learn high-order feature interactions explicitly. It can also be seen as performing the multi-head attention mechanism~\cite{vaswani2017attention} on a fully-connected graph.
	\item \textbf{Fi-GNN}~\cite{li2019fi}~(D)  models the features as a fully-connected graph and utilizes a gated graph neural network to model feature interactions.
	\item \textbf{InterHAt}~\cite{li2020interpretable}~(D) utilizes an attention mechanism to aggregate features, which are then multiplied by the original features to produce higher-order feature interactions.
\end{itemize}

\subsubsection{Implementation Details}
We implement the method using TensorFlow~\cite{abadi2016tensorflow} and Pytorch~\cite{paszke2019pytorch}.
The feature embedding size is set as 16 for all methods.
For a fair comparison, we set three layers for AutoInt, FiGNN, and GraphFM. There are two attention heads in AutoInt and GraphFM.
The implementation of the other compared baselines follows~\cite{song2019autoint} and ~\cite{tian2023directed}.
The optimal hyper-parameters are found via a grid search.
$\left [m_1, m_2, m_3\right ]$ are set as $\left [ 39, 20, 5 \right ]$, $\left [ 23, 10, 2 \right ]$, and $\left [ 7, 4, 2 \right ]$ for the Criteo, Avazu and MovieLens-1M datasets respectively.
We use Adam~\cite{kingma2014adam} to optimize all these models. 
The experiments were conducted over a server equipped with 8 NVIDIA Titan X GPUs.

\begin{table*}[t]\small
\centering\caption{Performance comparison of different methods on three datasets. The four model classes (A, B, C, D) are defined in Section~\ref{sect:baseline}. The last two columns are average improvements of our proposed model GraphFM compared with corresponding base models (``+'': increase, ``-'': decrease).
 We highlight the best performances on each dataset. Further analysis is provided in Section~\ref{sect:result}.}\label{tab:results}
\tabcolsep=0.28cm
\begin{tabular}{clcccccc||cc}
\toprule
\multicolumn{1}{c}{\multirow{2}{*}{Model}} & \multicolumn{2}{c}{Criteo} & \multicolumn{2}{c}{Avazu} & \multicolumn{2}{c}{MovieLens-1M}\\
\\\multicolumn{1}{c}{}
& AUC & LogLoss & AUC & LogLoss & AUC & LogLoss 
\\\midrule
 LR & 0.7820 & 0.4695 & 0.7560 & 0.3964 & 0.7716 & 0.4424\\
 \midrule
FM~\cite{rendle2010factorization} & 0.7836 & 0.4700 & 0.7706 & 0.3856 & 0.8252 & 0.3998\\  AFM\cite{xiao2017attentional} & 0.7938 & 0.4584 & 0.7718 & 0.3854 & 0.8227 & 0.4048\\
AFN~\cite{cheng2020adaptive} &0.8079 &0.4433 & 0.7786 & 0.3799 & 0.8771 & 0.4721\\
FmFM~\cite{sun2021fm2} & 0.8083 &0.4434 & 0.7746 & 0.3859 & 0.8821 & 0.3279\\
\midrule
NFM~\cite{he2017neural2} & 0.7957 & 0.4562 & 0.7708 & 0.3864 & 0.8357 & 0.3883\\
HOFM~\cite{blondel2016higher} & 0.8005 & 0.4508 & 0.7701 & 0.3854 & 0.8304 & 0.4013\\
DeepCrossing~\cite{shan2016deep} & 0.8009 &0.4513 & 0.7643 & 0.3889 & 0.8448 & 0.3814\\
CrossNet~\cite{wang2017deep} & 0.7907 & 0.4591 & 0.7667 & 0.3868 & 0.7968 & 0.4266\\
xDeepFM~\cite{lian2018xdeepfm} & 0.8009 & 0.4517 & 0.7758 & 0.3829 & 0.8286 & 0.4108\\
DCNV2~\cite{wang2021dcn} &0.8074 &0.4436 & 0.7666 & 0.3865 & 0.8833 & 0.4885\\
\midrule
AutoInt~\cite{song2019autoint} & 0.8084 & 0.4427 & 0.7781 & 0.3795 & 0.8823 & 0.3463\\
Fi-GNN~\cite{li2019fi} & 0.8077 & 0.4413 & 0.7778 & 0.3811 & 0.8792 & 0.3537\\
InterHAt~\cite{li2020interpretable} &0.8076 &0.4446 &0.7758 &0.3860	&0.8769 &0.3591\\
GraphFM (ours) & \textbf{0.8091} & \textbf{0.4399} & \textbf{0.7798} & \textbf{0.3781} & \textbf{0.8902} & \textbf{0.3259}\\
\bottomrule
\end{tabular}
\end{table*}

\subsection{Model Comparison}\label{sect:result}
\subsubsection{Evaluation of Effectiveness}
The performance comparison of these methods on three datasets is presented in Table \ref{tab:results}, from which we have the following observations.
Our proposed GraphFM achieves the best performance among all four classes of methods on three datasets. The performance improvement of GraphFM compared with the three classes of methods (A, B, C) is especially significant, above the $\mathbf{0.01}$-level. The aggregation-based methods including InterHAt, AutoInt, Fi-GNN and our GraphFM consistently outperform the other three classes of models, which demonstrates the strength of the aggregation strategy in capturing high-order relations. Compared with the strong aggregation-based baselines AutoInt and Fi-GNN, GraphFM still achieves large improvements in performance, especially on the MovieLens-1M dataset. The performance improvement on the other two datasets is also at the $\mathbf{0.001}$-level, which can be regarded as significant for the CTR prediction task~\cite{cheng2016wide,guo2017deepfm,song2019autoint,zhang2022novel,wu2020tfnet,xu2021disentangled}.
This improvement can be attributed to its combination with FM, which introduces feature interaction operations, and the interaction selection mechanism, which selects and models only the beneficial feature interactions. GraphFM outperforms the compared baselines by the largest margin on the MovieLens-1M dataset, whose feature size is the smallest among the three datasets. This is likely because the feature embedding size is not large enough for the other two datasets.

\begin{figure*}[t]
\centering
\subfigure[Criteo]{
\label{fig:criteo_auc} 
\begin{minipage}[b]{0.5\textwidth}
\includegraphics[width=1\textwidth]{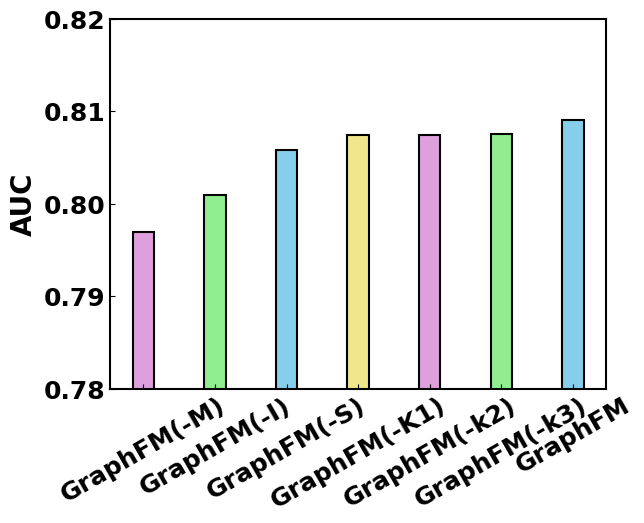}
\end{minipage}%
\begin{minipage}[b]{0.5\textwidth}
\includegraphics[width=1\textwidth]{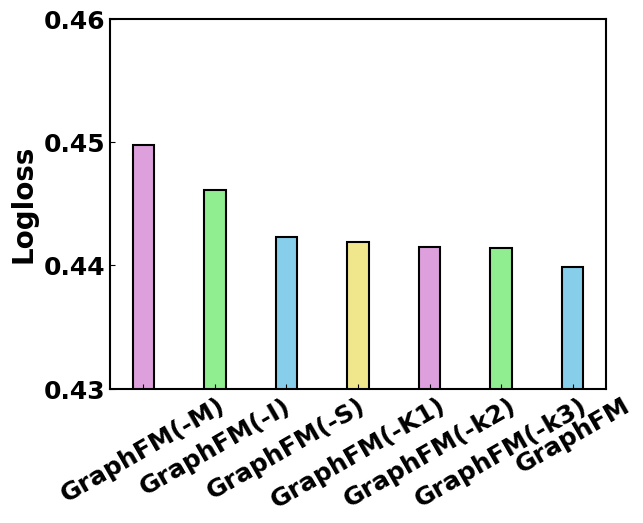}
\end{minipage}%
}

\subfigure[MovieLens-1M]{
\label{fig:movielens_auc} 
\begin{minipage}[b]{0.5\textwidth}
\includegraphics[width=1\textwidth]{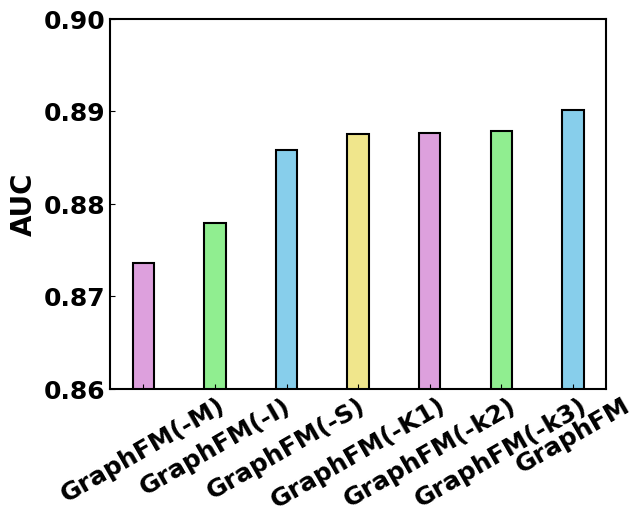}
\end{minipage}%
\begin{minipage}[b]{0.5\textwidth}
\includegraphics[width=1\textwidth]{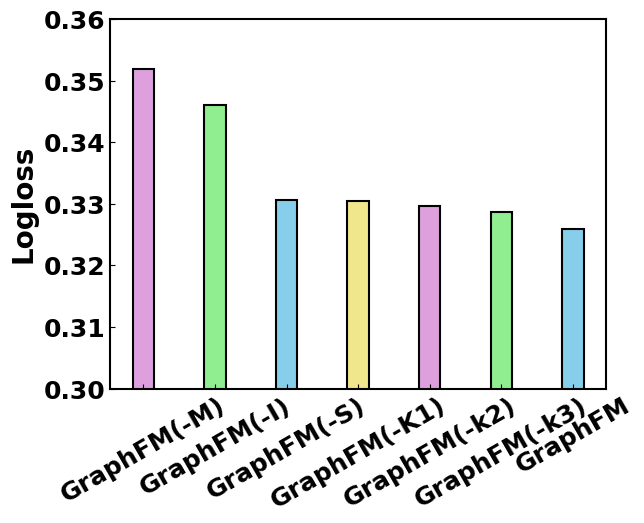}
\end{minipage}%
}%
\caption{Performance comparison of GraphFM with different components on the Criteo and MovieLens-1M datasets. Further analysis is provided in Section~\ref{sect:ablation}.}
\label{fig:ablation_study}
\end{figure*}

\subsection{Ablation Studies}\label{sect:ablation}
To validate the effectiveness of each component in GraphFM, we conduct ablation studies and compare several variants of it:
\begin{itemize}
\item \textbf{GraphFM(-S)}: \emph{interaction selection} is the first component in each layer of GraphFM, which selects only the beneficial feature interactions and treats them as edges. As a consequence, we can model only these beneficial interactions with the next \emph{interaction aggregation} component. To check the necessity of this component, we remove these components, so that all pairs of feature interactions are modelled as a fully-connected graph.
\item \textbf{GraphFM(-I)}: In the \emph{interaction aggregation} component, we aggregate the feature interactions instead of the neighbors' features, as in the standard GNNs. To check its rationality, we test the performance of directly aggregating neighborhood features instead of the feature interactions with them.
\item \textbf{GraphFM(-M)}: In the \emph{interaction aggregation} component, we use a multi-head attention mechanism to learn the diversified polysemy of feature interactions in different semantic subspaces. To check its rationality, we use only one attention head when aggregating.
\item \textbf{GraphFM(-$K_i$)}: Before obtaining the final representation of the feature, we concatenate and average the feature representation $e_k$ output from layer $k$. To study the degree of contribution of the features learned at different layers to the results, we use the feature representation of the $k$-th layer for direct prediction, and there are $K=3$ layers.
\end{itemize}

The performances of GraphFM and these four variants are shown in Fig.~\ref{fig:ablation_study}.
We observe that GraphFM outperforms all the ablative methods, which proves the necessity of all these components in our model.
The performance of GraphFM(-M) suffers from a sharp decrease compared with that of GraphFM, proving that it is necessary to transform and aggregate the feature interactions in multiple semantic subspaces to accommodate polysemy. 
Note that although we did not present the statistics here, we also tested the influence of the number of attention heads $H$. The performance of using only one head, i.e., GraphFM(-M), is worse than that of using two, and more attention heads do not lead to improvement in performance but introduce much greater time and space complexity.
GraphFM(-I) does not perform well either. This is reasonable, as without the interaction between features, neighborhood aggregation operation will only make neighboring features similar.
As a consequence, no feature interactions are guaranteed to be captured.
This interaction is also the most significant difference between GraphFM and GNN, and the resulting difference in terms of performance indicates that GraphFM is able to leverage the strength of FM to overcome the drawbacks of GNN in modelling feature interactions.
GraphFM(-S) achieves slightly worse performance than GraphFM, demonstrating that selecting and modelling only the beneficial interactions instead of all of them can avoid noise and make it easier to train the model.
The GraphFM(-$K_i$) results show that each layer can learn features that benefit the results. Although different features can have a positive effect on task prediction, the difference in effect is not large, and the prediction results are worse than those obtained by combining all features, demonstrating the need to merge the features of the $k$-th layers.

\begin{figure*}[p]
\centering
\subfigure[Criteo]{
\begin{minipage}[t]{0.6\linewidth}
\centering
\includegraphics[width=1\textwidth]{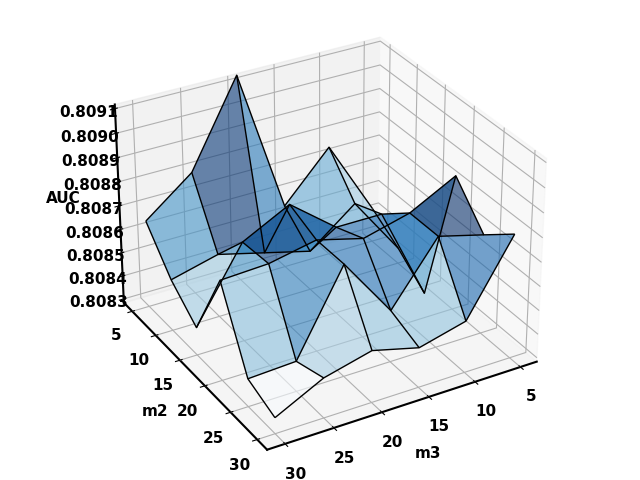}
\end{minipage}%
}%

\subfigure[Avazu]{
\begin{minipage}[t]{0.6\linewidth}
\centering
\includegraphics[width=1\textwidth]{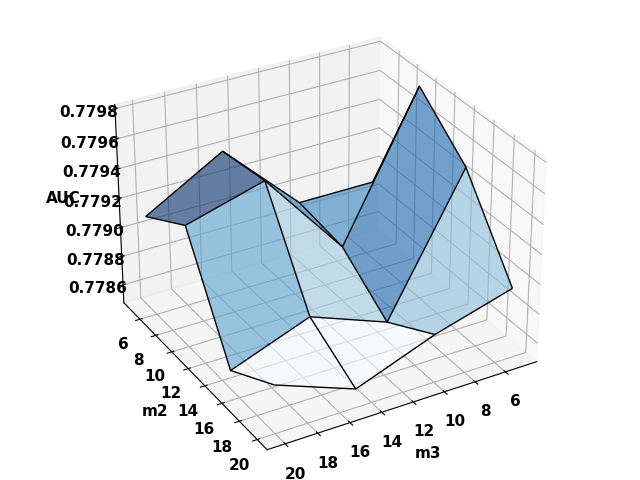}
\end{minipage}%
}%

\subfigure[MovieLens]{
\begin{minipage}[t]{0.6\linewidth}
\centering
\includegraphics[width=1\textwidth]{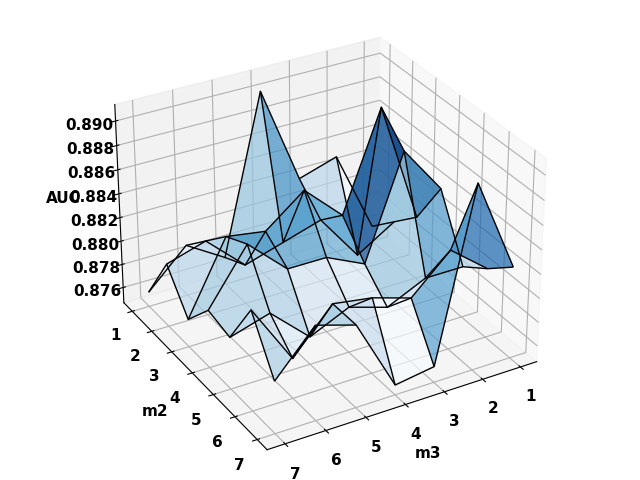}
\end{minipage}
}%
\centering
\caption{Model performance with respect to the size of the sampled neighborhood, where the ``neighborhood sample size'' refers to the number of neighbors sampled at each depth for $K = 3$ with $m_1 = n$, and $m_2, m_3$ with varying values.}
\label{fig:edge} 
\end{figure*}

\subsection{Study of Neighborhood Sampled Size}
The number of selected neighbors for each feature node is an important hyper-parameter that controls the number of features with which each feature interacts.
We thus investigate how the neighborhood sample size affects the model performance. 
As the total search space is too large, we only show the performance of our model with $K = 3, m_1=n$ and varying values of $m_2$ and $m_3$. 
The results on three datasets are summarized in Fig.~\ref{fig:edge}.

On the Criteo dataset, there are a total of 39 feature fields. We found that our model achieves the best performance with $m_1=39$, and $m_2 \times m_3=100$. The performances vary in the range of [0.8084, 0.8091], which proves that our model is quite robust, and not very sensitive to the size of the neighborhood sampled.
On the Avazu dataset, the model performance peaks with $m_1=23, m_2=10, m_3=2$ or $m_2=15, m_3=4$.
On the MovieLens-1M dataset, the model performance peaks when $m_2 \times m_3$ is approximately 9. We also found a diminishing trend for sampling larger or smaller neighborhoods.
In other words, the optimal neighborhood sample size depends on the dataset size.

\begin{figure*}[t]
\centering
\subfigure[label=1, pred=0.9736]{
\label{fig:layer} 
\begin{minipage}[b]{0.45\textwidth}
\includegraphics[width=1\textwidth]{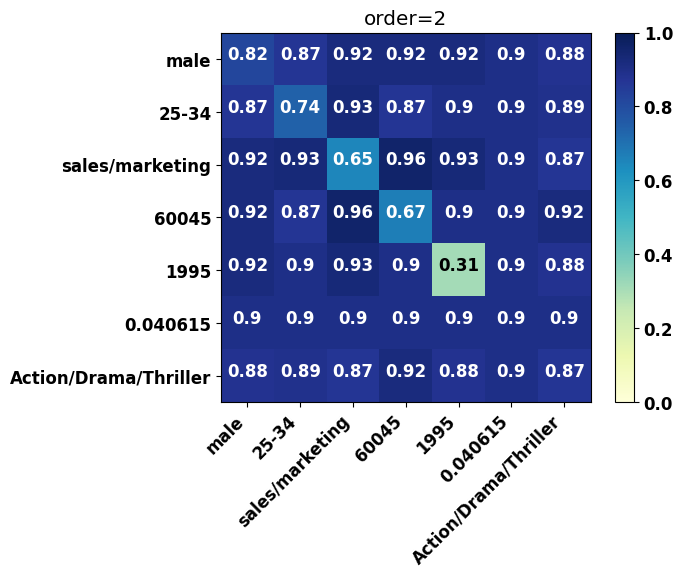} \\
\includegraphics[width=1\textwidth]{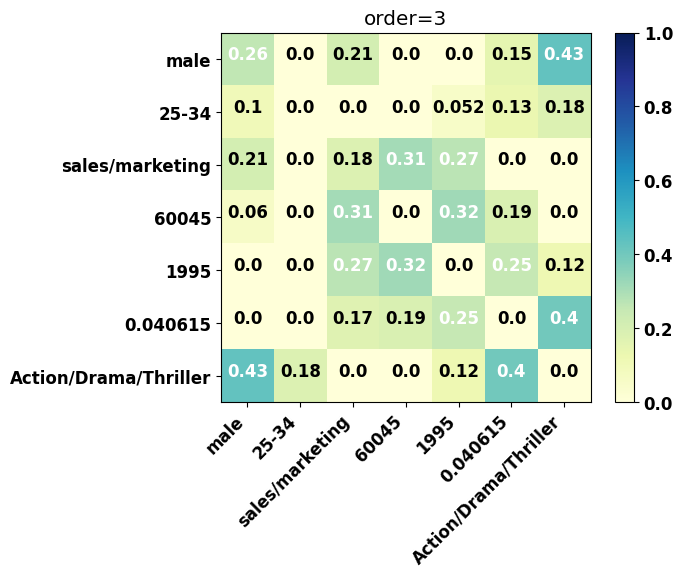} \\
\includegraphics[width=1\textwidth]{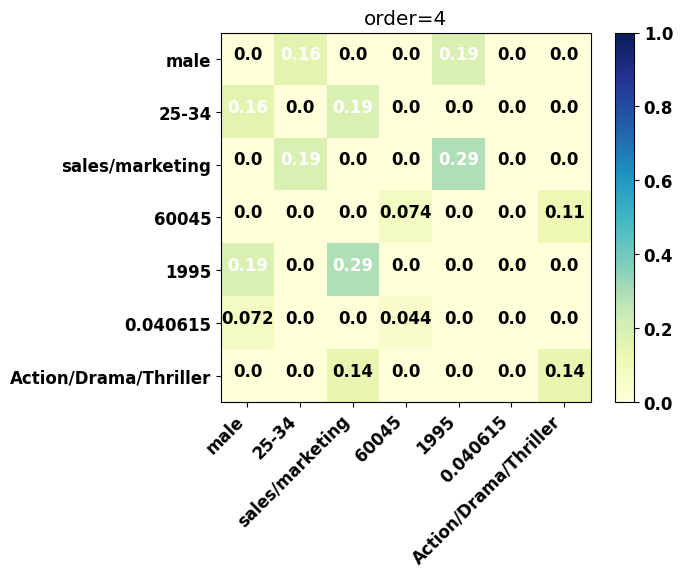}
\end{minipage}%
}
\subfigure[label=1, pred=0.0330]{
\label{fig:layer} 
\begin{minipage}[b]{0.45\textwidth}
\includegraphics[width=1\textwidth]{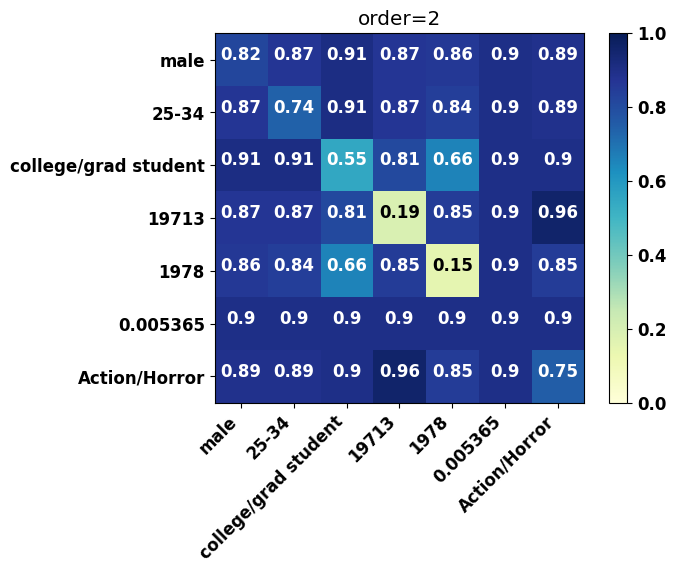} \\
\includegraphics[width=1\textwidth]{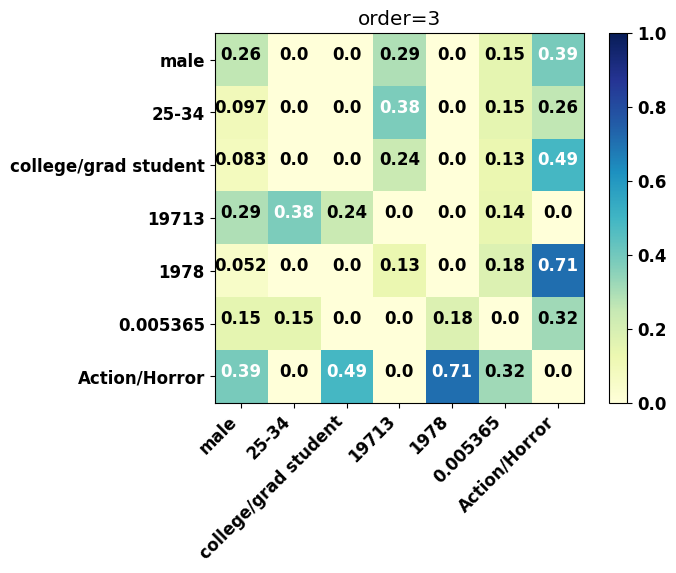} \\
\includegraphics[width=1\textwidth]{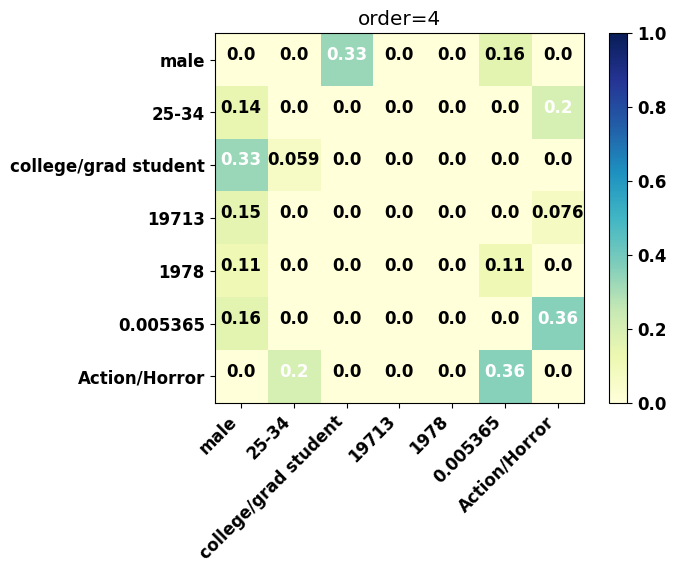}
\end{minipage}%
}
\caption{Heat maps of the estimated edge weights of correctly predicted instances (a) and incorrectly predicted instances (b) on the MovieLens-1M dataset, where positive edge weights indicate beneficial feature interactions. The axes represent feature fields (\textit{Gender}, \textit{Age}, \textit{Occupation}, \textit{Zipcode}, \textit{ReleaseTime}, \textit{WatchTime}, \textit{Genre}). }
\label{fig:vis}
\end{figure*}

\subsection{Visualization of the Interaction Graph}
Since the features along with selected beneficial feature interactions are treated as a graph, they can provide human readable interpretations of the prediction. Here, we visualize heat maps of the estimated edge weights of two cherry-pick instances on the MovieLens-1M dataset in Fig.~\ref{fig:vis}. We show the measured edge weights of each instance in the three layers, which select the order-2, order-3, and order-4 feature interactions. The positive edge weights indicate how beneficial feature interactions are. We set $S_1=7$, $S_2=4$, and $S_3=2$, which means that we only retain 7, 4, and 2 pairs of beneficial order-2, order-3, and order-4 feature interactions respectively. Therefore, there are only 7, 4, and 2 interaction feature fields for each feature field in each row for heat maps of order-2, order-3, and order-4, respectively. The axes represent feature fields (\textit{Gender}, \textit{Age}, \textit{Occupation}, \textit{Zipcode}, \textit{ReleaseTime}, \textit{WatchTime}, \textit{Genre}). \textit{Gender}, \textit{Age}, \textit{Occupation} and \textit{Zipcode} are users' demographic information. Note that \textit{Zipcode} indicates the users' place of residence. textit{ReleaseTime} and \textit{Gender} are the movie information. \textit{WatchTime} (\textit{Timestamp}) represents the time when users watched the movies.

From the two instances in Fig.~\ref{fig:vis}, we can obtain the following interesting observations.
We find that in the first layer, which models the second order feature interactions, these feature fields are difficult to distinguish when selecting the beneficial interactions. This suggests that almost all the second-order feature interactions are useful, which is why we sample all of them in the first layer, i.e., $m_1 = n$, except that the diagonal elements have the smallest values, which suggests that our designed interaction selection mechanism can classify the redundant self-interacting feature interactions, even though we keep and model all pairs of feature interactions.
The selected feature interactions of order-3 and order-4 mostly do not overlap in the correctly predicted instance (a). In instance (a), our model selects relevant feature fields (\textit{Gender}, \textit{Age}, \textit{ReleaseTime}, and \textit{WatchTime}) for \textit{Genre} in order-3 but selects the other two feature fields (\textit{Occupation} and \textit{Gender}) in order-4. 
However, in the wrongly predicted instances (b), the feature interactions of order-3 and order-4 mostly do not overlap. 

This proves that our model can indeed select meaningful feature combinations and model feature interactions of increasing orders with multiple layers in most cases, rather than selecting the redundant feature combinations of 
the same feature fields.
We can also find some meaningful feature combinations in common cases. For example, \textit{Gender} is usually relevant to the feature fields \textit{Age}, \textit{occupation}, and \textit{WatchTime}, while \textit{Age} is usually relevant to the feature fields \textit{Gender}, \textit{WatchTime}, and \textit{Genre}. This provides some rationale for the model prediction.

\section{Conclusion and Future Work}
In this work, we disclose the relationship between FMs and GNNs, and seamlessly combine them to propose a novel model GraphFM for feature interaction learning.
The proposed model leverages the strengths of FMs and GNNs and solves their respective drawbacks.
At each layer of GraphFM, we select the beneficial feature interactions and treat them as edges in a graph. Then, we utilize a neighborhood/interaction aggregation operation to encode the interactions into feature representations.
By design, the highest order of feature interaction increases at each layer and is determined by layer depth; thus, the feature interactions of order up to the highest can be learned.
GraphFM models high-order feature interactions in an explicit manner, and can generate human readable explanations of outcomes.  
The experimental results show that GraphFM outperforms the state-of-the-art baselines by a large margin.
In addition, we conduct extensive experiments to analyse how the highest order of feature interactions and the number of modelled feature interactions influence model performance, which can help us gain deeper insight into feature interaction modelling.
In the future, we aim to investigate whether the proposed method can also benefit graph representation learning, and graph/node classification tasks.

\bmhead{Acknowledgements}

We would like to thank the anonymous reviewers for their valuable comments and suggestions, allowing us to improve the quality of this paper. This work is sponsored by the National Science Foundation of China (62141608).

\bibliography{mybibliography}

\par\noindent 
\parbox[t]{\linewidth}{
\noindent\parpic{\includegraphics[height=1.5in,width=1in,clip,keepaspectratio]{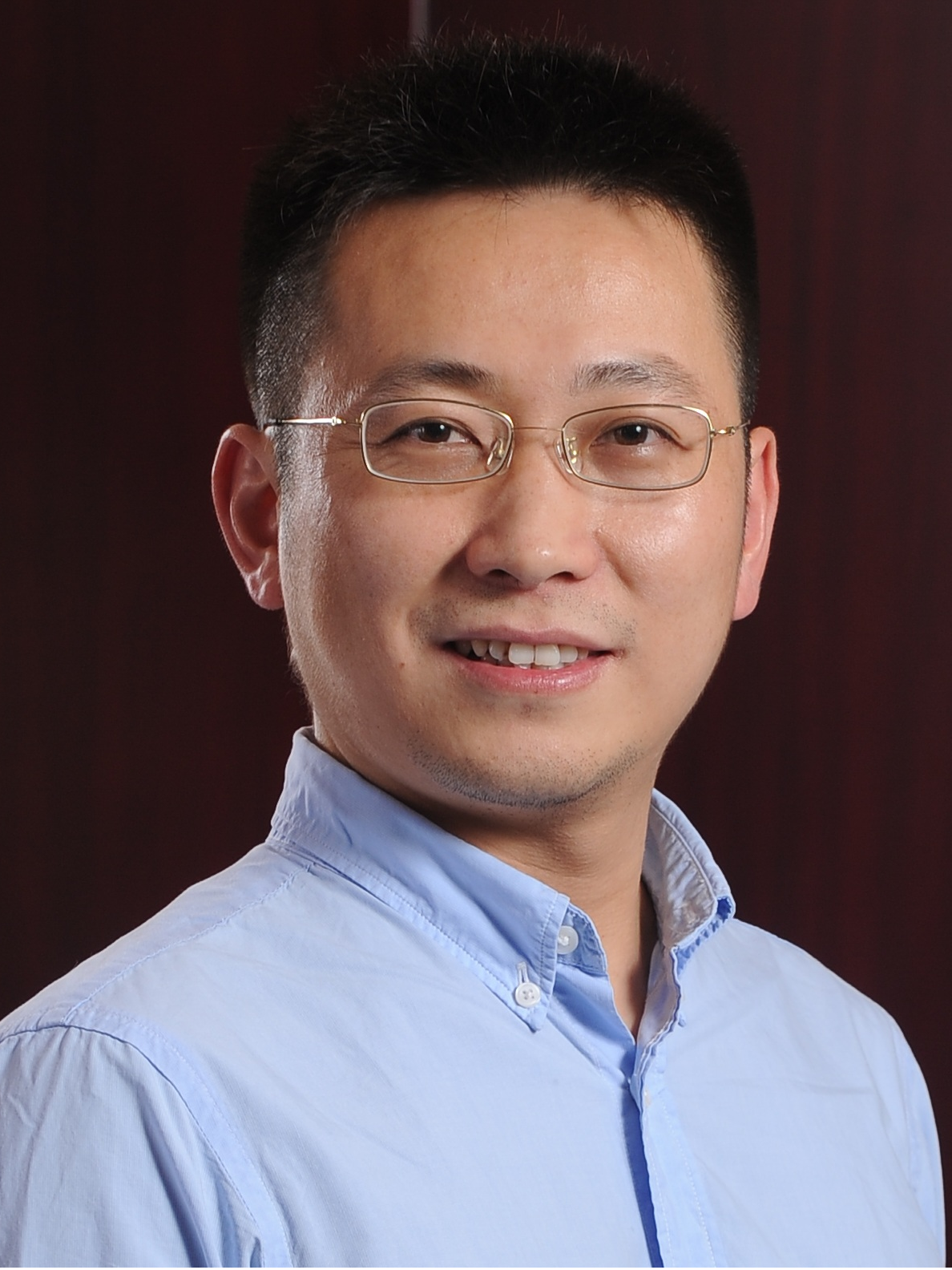}}
\noindent {\bf Shu Wu}\
received his B.S. degree from Hunan University, China, in 2004, his M.S. degree from Xiamen University, China, in 2007, and his Ph.D.degree from the University of Sherbrooke, Quebec, Canada. He is an Associate Professor at the NLPR, CASIA.

His research interests include data mining and pattern recognition.

E-mail: shu.wu@nlpr.ia.ac.cn

ORCID iD: 0000-0003-2164-3577}
\vspace{1\baselineskip}

\par\noindent 
\parbox[t]{\linewidth}{
\noindent\parpic{\includegraphics[height=1.5in,width=1in,clip,keepaspectratio]{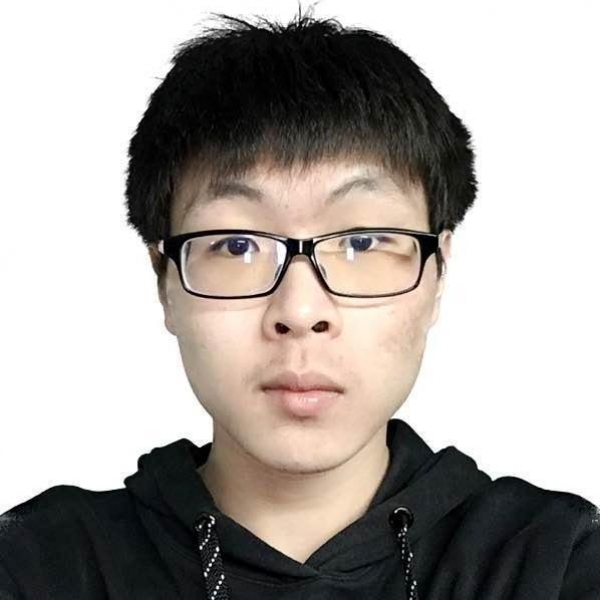}}
\noindent {\bf Zekun Li}\
is a Ph.D. at the University of California, Santa Barbara. He obtained a master's degree from the University of Chinese Academy of Sciences, Beijing, China, in 2021, and the B.Eng degree from Shandong University, China, was obtained in 2018. 

His research interests include data mining, recommender systems, and natural language processing.

E-mail: zekunli@cs.ucsb.edu}
\vspace{1\baselineskip}

\par\noindent 
\parbox[t]{\linewidth}{
\noindent\parpic{\includegraphics[height=1.5in,width=1in,clip,keepaspectratio]{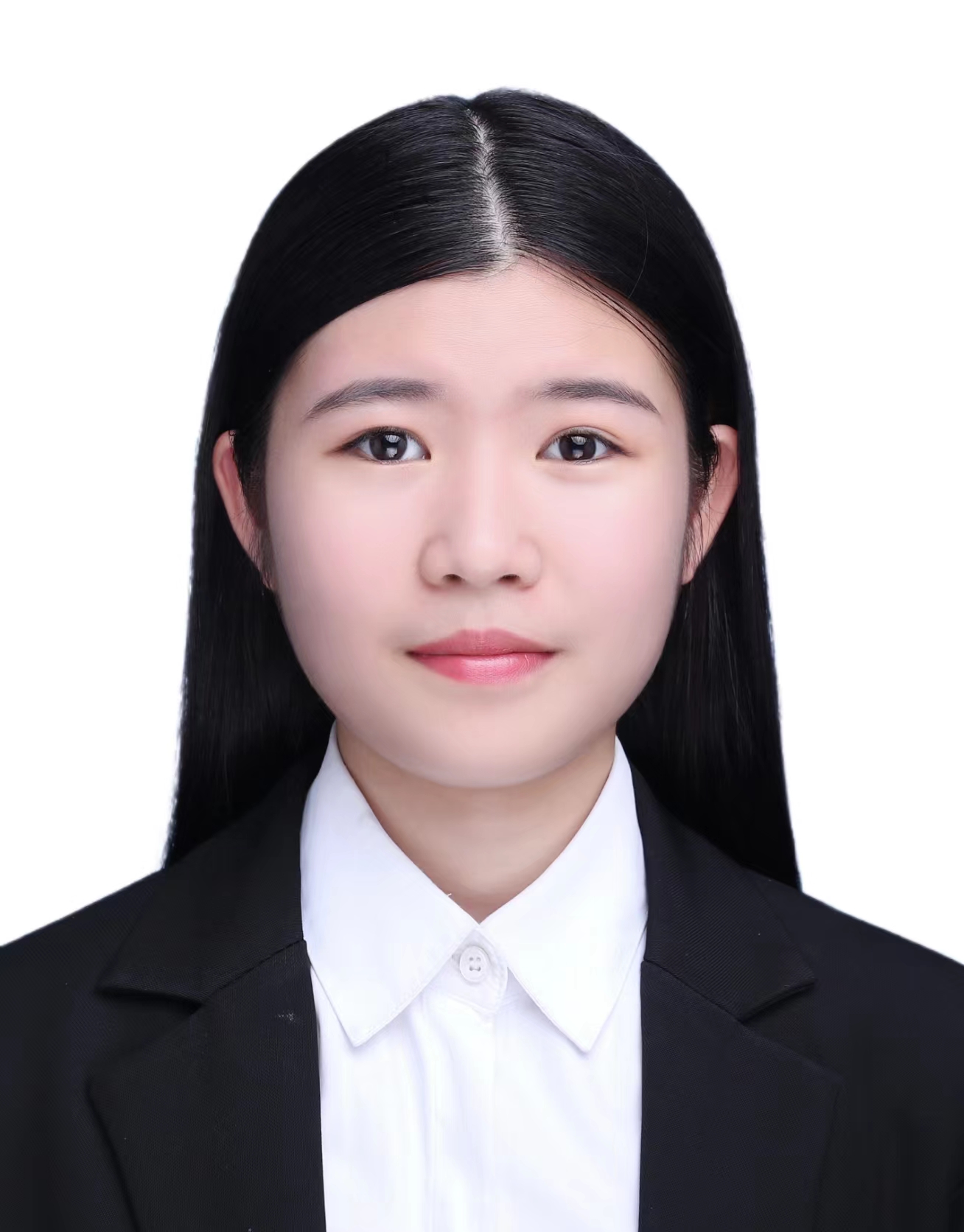}}
\noindent {\bf Yunyue Su}\
is with the Institute of Automation, Chinese Academy of Sciences. She received her B.S degree from the Computer Network Information Center, University of Chinese Academy of Sciences, China, in 2023. 

Her research interests include data mining, machine learning, and recommender systems.

E-mail: yunyue.su@ia.ac.cn}
\vspace{1\baselineskip}

\par\noindent 
\parbox[t]{\linewidth}{
\noindent\parpic{\includegraphics[height=1.5in,width=1in,clip,keepaspectratio]{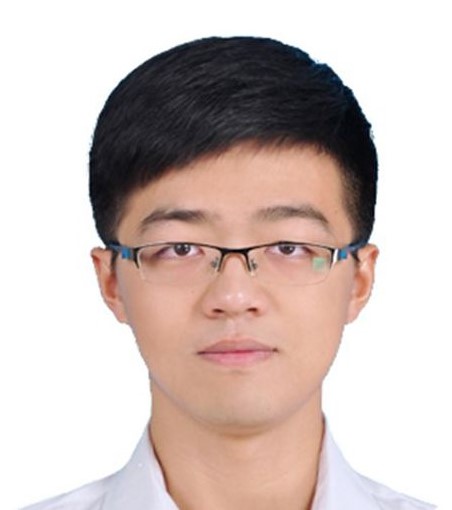}}
\noindent {\bf Zeyu Cui}\
is with the Alibaba Group, DAMO Institute. He obtained his Ph.D. degree from the School of Artificial Intelligence, University of the Chinese Academy of Sciences, Beijing, China. He received the B.S degree from North China Electric Power University, China, in 2016. 

His research interests include data mining, machine learning, and recommender systems.

E-mail: cuizeyu15@gmail.com}
\vspace{1\baselineskip}

\par\noindent 
\parbox[t]{\linewidth}{
\noindent\parpic{\includegraphics[height=1.5in,width=1in,clip,keepaspectratio]{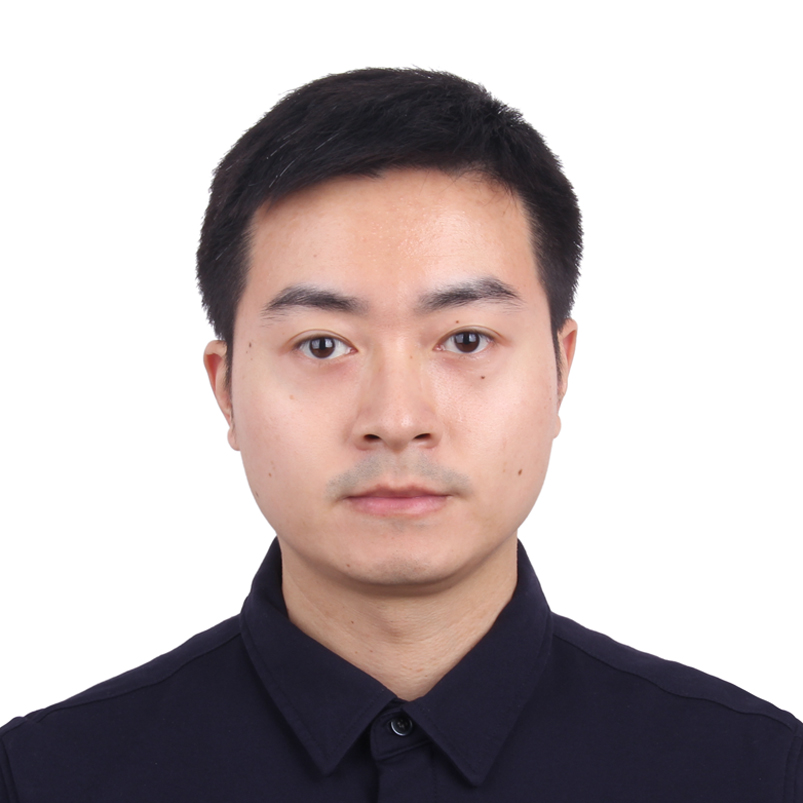}}
\noindent {\bf Xiaoyu Zhang}\
received a Ph.D. degree in pattern recognition and intelligent systems from the Institute of Automation, Chinese Academy of Sciences, Beijing, China, in 2010. He is currently a Professor with the Institute of Information Engineering, Chinese Academy of Sciences, Beijing, China. 

His research interests include artificial intelligence, data mining, computer vision, etc.

E-mail: zhxy333@gmail.com

ORCID iD: 0000-0001-5224-8647}
\vspace{1\baselineskip}

\par\noindent 
\parbox[t]{\linewidth}{
\noindent\parpic{\includegraphics[height=1.5in,width=1in,clip,keepaspectratio]{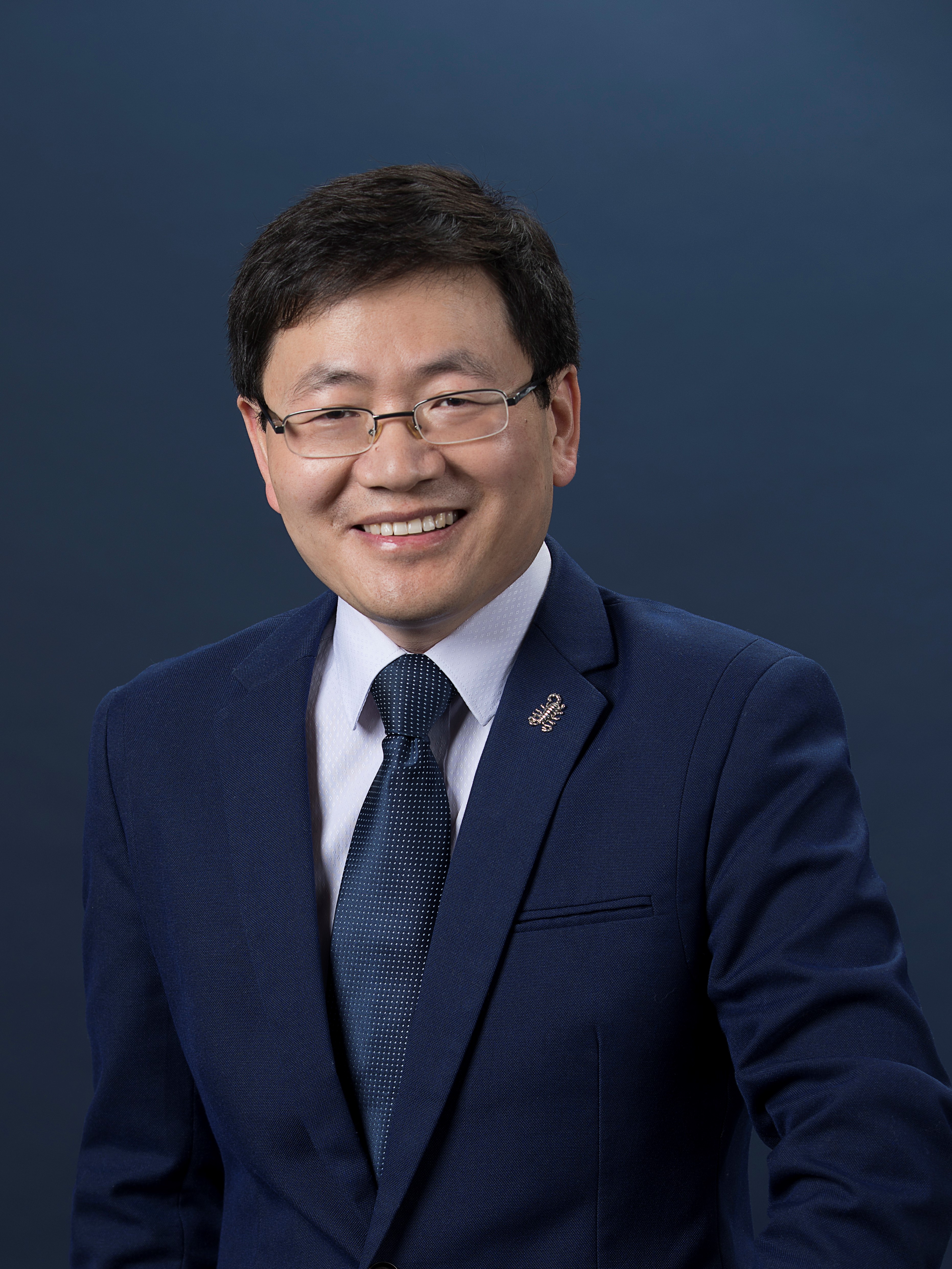}}
\noindent {\bf Liang Wang}\
received a Ph.D. degree from the National Laboratory of Pattern Recognition (NLPR), Institute of Automation, Chinese Academy of Sciences, Beijing, China, in 2004. He is currently a Professor at NLPR, CASIA. 

His major research interests include computer vision, pattern recognition, machine learning, and data mining. 

E-mail: liang.wang@nlpr.ia.ac.cn}
\vspace{1\baselineskip}

\end{document}